\documentclass{article} 
\usepackage[preprint]{colm2026_conference}

\usepackage{microtype}
\usepackage{hyperref}
\usepackage{url}
\usepackage{booktabs}
\usepackage{xspace}
\usepackage{graphicx}
\usepackage{xcolor}
\usepackage[table]{xcolor}
\usepackage[most]{tcolorbox}
\usepackage{enumitem}
\newtcolorbox{promptbox}{
  breakable,
  colback=gray!4,
  colframe=gray!60,
  boxrule=0.5pt,
  arc=1.5pt,
  left=6pt,
  right=6pt,
  top=6pt,
  bottom=6pt,
  before skip=8pt,
  after skip=8pt
}
\usepackage{array}
\usepackage{listings}
\usepackage{placeins}

\usepackage[table]{xcolor}
\usepackage{tabularx}
\usepackage{array}

\usepackage{lineno}

\definecolor{darkblue}{rgb}{0, 0, 0.5}
\hypersetup{colorlinks=true, citecolor=darkblue, linkcolor=darkblue, urlcolor=darkblue}

\title{\model: Structured Autonomy for Agentic Gamified Learning in Cybersecurity}

\author{
    \textbf{Ivan Hornung\textsuperscript{1}},
    \textbf{Deepthi Marasinghe Arachchige\textsuperscript{1}}, 
    \textbf{Tharindu Kumarage\textsuperscript{2,\thanks{Work does not relate to the position at Amazon.}}}, 
    \\
    \textbf{Garima Agrawal\textsuperscript{1}},
    \textbf{Yuli Deng\textsuperscript{1}},
    \textbf{Ying-Chih Chen\textsuperscript{1}},
    \textbf{Huan Liu\textsuperscript{1}},
    \\
    \textsuperscript{1}Arizona State University,
    \textsuperscript{2}Amazon AGI
    \\
    \{ihornung, dsmarasi, garima.agrawal, ydeng19, ychen495, huanliu\}@asu.edu
}

\newcommand{\model}{\textbf{Cyber\textsc{agents}}\xspace}

\begin{document}

\ifcolmsubmission
\linenumbers
\fi

\maketitle

\begin{abstract}

Gamification is especially effective in learning domains requiring active problem-solving and iterative skill-building, such as cybersecurity education. Generative AI agents offer a path to delivering such experiences adaptively at scale, but introduce well-documented risks in educational settings: inconsistent behavior, hallucinated reasoning, and misalignment with pedagogical frameworks. Grounding these systems in learning science is therefore essential. We present \model, an agentic framework for gamified cybersecurity learning that enables structured autonomy through ontology-guided validation, schema-governed behavioral control, and competency-based progression. The system is organized around a competency-based progression model that structures topics by difficulty and prerequisite relationships, reflecting evidence-based principles of scaffolded instruction. The learning loop is decomposed into four specialized agents: challenge, support, evaluation, and reward, each governed by behavioral schemas that encode operational modes and progression logic, bounding agent autonomy without eliminating generative flexibility. A cybersecurity ontology validates all generated content prior to display, enforcing domain-consistent reasoning and safety constraints. We evaluate \model through classroom deployment with undergraduate students, complemented by expert evaluations from educators and domain specialists. Results indicate improved engagement, clearer feedback interpretation, and greater learner trust in AI-generated responses when behavioral schemas and ontology validation are active. Preliminary comparisons with an unconstrained configuration further support the role of structured control in stabilizing instructional behavior. These findings offer a blueprint for designing pedagogically grounded agentic gamified learning systems.

\end{abstract}

\section{Introduction}

Maintaining sustained engagement while ensuring reliable learning remains a central challenge in technical education. Gamification, through progressive challenges, adaptive instructional support, structured feedback loops, and reward progression, has been shown to enhance cognitive engagement and retention, particularly in practice-driven domains~\citep{dichev2017gamifying,sailer2020gamification,smiderle2020impact, gini2025role,nguyen2025does}. Cybersecurity education, however, presents additional complexity: concepts are abstract, adversarial, and high-stakes, and traditional lecture-driven instruction often fails to provide iterative reasoning practice or timely corrective feedback. As AI accelerates both offensive and defensive capabilities, preparing a skilled cybersecurity workforce requires learning environments that are interactive, scalable, and trustworthy~\citep{prakash2024artificial,lazarov2023interactive,grover2023cybersecurity, urias2017dynamic}

Recent advances in large language models (LLMs) and agentic AI systems enable dynamic, personalized learning experiences~\citep{wang2026large}. Generative models can produce challenges, explanations, and feedback on demand, but they also introduce instability: LLMs may hallucinate domain details, while unconstrained multi-agent systems can produce inconsistent guidance, conflicting feedback, or pedagogically misaligned behavior~\citep{alansari2025large,huang2025survey}. In educational settings, agents must generate correct domain content while respecting role boundaries, progression logic, and instructional dependencies. Without explicit control, dynamic prompting can lead to instructional drift~\citep{guo2024large, hammond2025multi}.

These challenges are amplified in gamified environments. Static designs with fixed prompts reduce risk but eliminate adaptivity. Fully dynamic systems preserve flexibility but risk incoherence and loss of trust. What is missing is a principled way to bound agentic autonomy, constraining both how agents behave and what domain relationships they may express, while preserving the benefits of generative interaction~\citep{borchers2025can,nazaretsky2025critical}.

We present \model, an ontology-guided agentic framework for
gamified cybersecurity education that addresses this gap through layered control mechanisms. First, a competency-mapping module organizes curriculum topics by difficulty, prerequisite structure, and conceptual clusters, providing structured initialization for gameplay. Second, the learning loop is decomposed into four specialized agents---Challenge, Buddy, Critic, and Reward---that reflect familiar roles in educational practice, including task setting, tutor-like guidance, instructor-style feedback, and motivational reinforcement. Each agent is governed by explicit behavioral schemas encoding operational modes, trigger conditions, progression logic, and evaluation criteria, introducing bounded autonomy: agents adapt to learner interaction while remaining within predefined pedagogical constraints. Third, all generated content is validated prior to display using a cybersecurity ontology~\citep{agrawal2023aiseckg} that enforces entity-type constraints, permissible domain relations, and safety rules ~\citep{agrawal2022building, agrawal2024can}. Together, these layers regulate both instructional behavior and domain semantics.

\model is deployed as a live, playable system. We evaluate it with 24 undergraduate students through classroom deployment, combining observational analysis, post-interaction surveys, and automated LLM-as-judge assessment. We further compare schema- and ontology-constrained agents against unconstrained baselines through preliminary ablation. Results show improved engagement, clearer feedback interpretation, stronger perceived trust in AI responses, and reduced domain inconsistencies when structured behavioral metadata and ontology validation are applied.

Our findings suggest that reliable AI-supported learning in high-stakes technical domains benefits from explicit control over both agent behavior and domain reasoning. By embedding pedagogical schemas and ontology constraints within a dynamic multi-agent architecture, \model offers a practical design blueprint for safe and adaptive agentic education systems.

\paragraph{Contributions.}
\begin{itemize}
    \item \textbf{Schema-governed agentic architecture:} We propose a layered multi-agent gamification framework in which agent behavior is bounded by explicit schemas and coordinated through an orchestrator, enabling structured control over dynamic interaction.

    \item \textbf{Competency-structured game initialization:} We design a competency-mapping module that organizes curriculum topics by difficulty, prerequisites, and concept clusters, providing principled progression for adaptive gameplay.

    \item \textbf{Ontology-constrained reasoning for safety and correctness:} We integrate domain-level validation through a cybersecurity ontology that enforces entity-type constraints and permissible relations, reducing hallucinations and unsafe instructional drift.

    \item \textbf{Empirical validation with preliminary ablation:} Through classroom deployment with undergraduate students, we show improvements in engagement, feedback clarity, trust, and domain consistency; preliminary comparisons with unconstrained baselines further support the role of structured control in stabilizing instructional behavior.
\end{itemize}




\section{Related Work}
Prior work in gamified learning and AI-enabled education provides a strong foundation for adaptive cybersecurity learning environments. \citet{suresh2024application} review gamification frameworks and show how AI can adapt rewards, challenge difficulty, and personalized scaffolding to support diverse learners. Their findings link AI-driven adaptation to core learning science principles such as feedback-centered learning, learner agency, and mastery progression. Similarly, \citet{kassenkhan2025gamification} identify AI-supported gamification as a promising approach for fostering higher-order skills such as critical thinking through real-time guidance and situational feedback. Complementing these perspectives, \citet{alenezi2023teacher} report that instructors view AI-assisted educational tools as strong motivators that improve student engagement and learning outcomes, particularly through timely automated feedback. Together, this literature shows that AI-enhanced gamification can strengthen motivation, engagement, and conceptual understanding.

At the same time, LLM-based tutoring systems have shown the potential of conversational agents for adaptive instruction, but they remain vulnerable to hallucinations and inconsistent feedback in structured technical domains such as cybersecurity~\citep{kasneci2023chatgpt, ji2023survey,bang2023multitask, rudolph2023chatgpt}. Other work has explored grounding LLM outputs with ontologies and knowledge graphs to improve factual consistency and interpretability~\citep{pan2024unifying, kang2023knowledge, agrawal2024cyberq, zhao2025ontology, zhao2025cyberbot}, while multi-agent LLM frameworks distribute responsibilities across role-specialized agents~\citep{wu2024autogen, park2023generative, hong2308metagpt}. However, these directions have largely evolved independently: gamified learning systems often lack domain-grounded reasoning, knowledge-grounded LLM methods are not designed for interactive pedagogy, and multi-agent architectures rarely align agent roles with competency-based cybersecurity learning. \model addresses this gap through a unified framework that combines gamified interaction, ontology-guided validation, and coordinated pedagogical agents for cybersecurity education.

\section{CyberAgents Framework}

\begin{figure*}[t]
    \centering
    \includegraphics[width=0.9\textwidth]{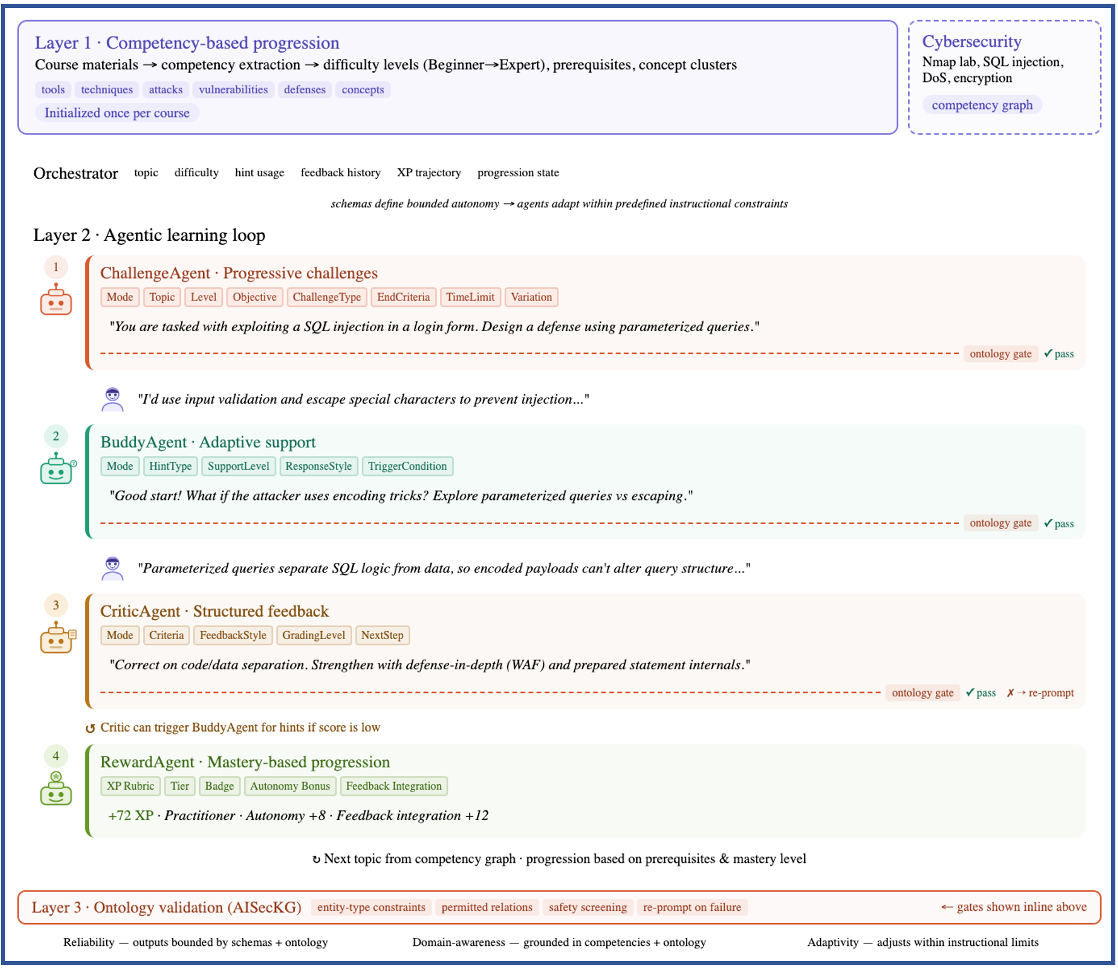}
    \caption{\textbf{CyberAGENTS system architecture.} Layer 1 initializes competency structure once per course. Layer 2 orchestrates the four-agent dialogue loop (challenge → response → evaluation → reward) with the learner; each agent operates under predefined behavioral schemas that bound generative autonomy. Layer 3 validates all agent outputs against the AISecKG ontology before display, re-prompting on failure.}
    \label{fig:architecture}
\end{figure*}

\model is a layered, ontology-guided multi-agent framework for agentic gamified learning in cybersecurity. It is designed to support adaptive, practice-driven instruction while preserving pedagogical alignment, domain correctness, and progression coherence. The framework enables \textit{structured autonomy} through three complementary control components: competency-based progression, schema-governed behavioral control, and ontology-guided validation. Figure~\ref{fig:architecture} presents the overall system design.

Cybersecurity learning particularly benefits from this structure. Unlike domains centered on memorization or static concept recall, cybersecurity requires learners to reason about attacks, defenses, vulnerabilities, tools, and system interactions in an adversarial and high-stakes environment. Effective gamification in this setting therefore requires more than superficial points or badges: it must align progressive challenge design, adaptive support, structured feedback, and reward progression with meaningful instructional goals. At the same time, agentic AI systems introduce additional risks. Generative agents can create dynamic and personalized interactions, but without explicit control they may drift from pedagogical goals, provide inconsistent support, or generate domain-inaccurate content. \model addresses this tension through a layered design that preserves generative flexibility while bounding agent behavior and domain reasoning through structured autonomy.

\subsection{Competency-Based Progression}

\model begins by extracting structured competencies from course materials using a cybersecurity-specific competency schema. The extractor identifies key entities such as tools, techniques, concepts, attacks, vulnerabilities, and defenses; assigns difficulty levels from Beginner to Expert; and maps instructional relations relevant to gameplay design, including prerequisites, concept clusters, and learning enablers. The resulting competency graph encodes topical coverage and progression structure, serving as the initialization blueprint for challenge generation.

This competency-based progression ensures that gameplay is curriculum-aligned from the outset. Rather than sampling arbitrary prompts, the system draws from a leveled representation of domain knowledge, allowing challenge progression to follow explicit instructional structure. This is particularly important in cybersecurity education, where conceptual dependencies and difficulty progression must be carefully managed to avoid cognitive overload or fragmented learning.

\subsection{Agentic Learning Loop}

Gamification in \model is operationalized through a structured multi-agent learning loop. Instead of relying on a single generative model to improvise across multiple instructional functions, the framework decomposes gameplay into four specialized agents: \textit{ChallengeAgent}, \textit{BuddyAgent}, \textit{CriticAgent}, and \textit{RewardAgent}. Together, these agents enact the cycle of challenge $\rightarrow$ learner attempt $\rightarrow$ evaluation $\rightarrow$ progression.

At runtime, a central orchestrator coordinates these agents and maintains shared state across turns, including topic, difficulty, hint usage, feedback history, and XP trajectory. This shared state allows the system to preserve coherent progression across interactions while adapting support and challenge difficulty to learner behavior. The \textit{ChallengeAgent} generates competency-aligned tasks with calibrated difficulty and explicit objectives. The \textit{BuddyAgent} provides adaptive instructional support based on learner performance and interaction signals. The \textit{CriticAgent} evaluates learner responses and produces formative feedback. The \textit{RewardAgent} assigns XP, badges, and progression indicators according to a mastery-oriented reward rubric.

This role-based decomposition supports the core gamified learning functions of progressive challenges, adaptive support, structured feedback, and mastery-based rewards. More importantly, it enables modular control over instructional dynamics, making the learning loop more interpretable and stable than monolithic prompting approaches.

\subsection{Schema-Governed Behavioral Control}

To ensure that adaptive agent behavior remains instructionally coherent, \model defines explicit behavioral schemas for each agent. These schemas function as structured metadata contracts specifying how an agent operates, when it intervenes, and what form its outputs may take. Agent behavior is parameterized through predefined fields governing operational mode, difficulty level, trigger conditions, response style, evaluation criteria, and reward logic.

These schemas implement \textit{structured autonomy}: agents retain generative flexibility within defined bounds, but cannot deviate from prescribed instructional roles. Challenge generation is constrained by difficulty levels and explicit objectives; instructional support is regulated through trigger conditions and assistance levels; evaluation follows structured criteria; and rewards are tied to an experience-based mastery rubric. This prevents role drift while preserving adaptivity across learners.

Schemas also support cross-agent coordination. Because agents expose structured properties such as difficulty level, hint usage, grading mode, and XP tier, the orchestrator can maintain coherent progression across turns. Adjustments in challenge complexity, instructional support, and reward allocation are governed by shared state signals rather than ad hoc prompting alone. In this way, schema-governed behavioral control regulates instructional conduct and ensures that the gamified interaction remains aligned with pedagogical goals.

\subsection{Ontology-Guided Validation}
While schemas regulate instructional behavior, domain correctness is enforced through ontology-guided validation. \model integrates AISecKG~\cite{agrawal2023aiseckg}, a cybersecurity ontology that defines valid entity types (e.g., tool, technique, attack, vulnerability, defense, system) and permissible relations among them (e.g., exploits, detects, counters, uses, can\_harm). This ontology acts as a reasoning constraint layer applied to agent-generated content before learner exposure.

During gameplay, outputs from the \textit{ChallengeAgent}, \textit{BuddyAgent}, and \textit{CriticAgent} are checked against ontology-defined entity categories, relation patterns, and unsafe-content rules. Content that violates semantic constraints or safety thresholds is flagged and re-prompted under stricter conditions before presentation. This pre-display gating complements schema-governed behavioral control: schemas constrain pedagogical behavior and progression logic, while ontology-guided validation constrains cybersecurity reasoning and semantic validity. Together, these layers regulate both instructional conduct and domain semantics in a dynamically generated learning environment.

\subsection{Agent Roles and Gamified Learning Functions}

Each agent in \model operationalizes a distinct gamified learning function under explicit schema and ontology constraints. This mapping allows the framework to translate high-level motivational and instructional principles into modular agent behavior.

\paragraph{ChallengeAgent (Progressive Challenges).}
The \textit{ChallengeAgent} is responsible for generating competency-aligned tasks derived from the structured competency graph. Its behavior is parameterized through schema fields such as \textit{Mode}, \textit{Topic}, \textit{Level}, \textit{Objective}, \textit{ChallengeType}, \textit{EndCriteria}, \textit{TimeLimit}, and \textit{Variation}. These properties encode progression logic, calibrate difficulty, and preserve clarity of learning goals. Controlled diversity is introduced through mode and variation fields, while ontology-guided validation ensures that challenge content remains domain-consistent before presentation.

\paragraph{BuddyAgent (Adaptive Support).}
The \textit{BuddyAgent} provides adaptive support through structured properties including \textit{Mode} (on-demand, proactive, collaborative), \textit{HintType}, \textit{SupportLevel}, \textit{ResponseStyle}, and \textit{TriggerCondition}. These parameters regulate when and how support is provided. For example, proactive intervention may follow repeated incorrect attempts, while on-demand support preserves learner autonomy. Support levels enable gradual fading of assistance as learner competence increases, preventing over-assistance while maintaining responsiveness to learner needs.

\paragraph{CriticAgent (Structured Feedback).}
The \textit{CriticAgent} evaluates learner responses using schema-defined properties such as \textit{Mode}, \textit{Criteria}, \textit{FeedbackStyle}, \textit{GradingLevel}, and \textit{NextStep}. This transforms evaluation into a transparent and actionable feedback mechanism rather than an opaque generative judgment. Explicit criteria anchor assessment to defined instructional dimensions, while next-step options preserve forward progression. Ontology-guided validation further reduces hallucinated or unsafe explanatory content, ensuring that critique remains both pedagogically constructive and domain-aligned.

\paragraph{RewardAgent (Mastery-Based Progression).}
The \textit{RewardAgent} governs motivational continuity through an experience-based XP rubric. Rather than assigning static points, it computes progression signals using structured inputs such as challenge summary, hint usage, feedback integration, and reasoning accuracy. The agent maps performance to predefined tiers---\textit{Apprentice}, \textit{Explorer}, \textit{Practitioner}, \textit{Expert}, and \textit{Legendary}---with corresponding XP ranges and badge signals. This keeps reward progression aligned with mastery and metacognitive growth rather than raw correctness alone.

\paragraph{Design Implications.}
Taken together, these components illustrate a broader design principle for high-stakes agentic learning systems: generative flexibility should be bounded through structured autonomy rather than eliminated through static scripting. In \model, competency-based progression provides structured initialization, schema-governed behavioral control regulates instructional behavior, the orchestrator preserves coherent progression, and ontology-guided validation constrains domain reasoning and safety. This combination enables agentic gamified learning in cybersecurity while maintaining reliability, domain-awareness, and pedagogical alignment.

\paragraph{Mapping of Agents to Gamified Learning Functions.} Table~\ref{tab:pillar_mapping} summarizes the mapping between specialized agents and core gamified learning functions; representative ontology artifacts, prompt templates, and behavioral schema are provided in Appendix~\ref{app:model_schema}.

\begin{table}[h]
\centering
\begin{tabular}{ll}
\hline
\textbf{Agent} & \textbf{Gamified Learning Function} \\
\hline
ChallengeAgent & Progressive Challenges \\
BuddyAgent & Adaptive Support \\
CriticAgent & Structured Feedback \\
RewardAgent & Mastery-Based Progression \\
\hline
\end{tabular}
\caption{Mapping of \model to core gamified learning functions.}
\label{tab:pillar_mapping}
\end{table}

\section{Evaluation}

\subsection{System and Experimental Setup}

\model was implemented as a web-based interactive learning system with a React frontend and a Python Flask backend, and deployed as containerized services on Google Cloud Run. Firebase Authentication supported user sign-in, while Firestore was used to store user records, gameplay state, and interaction logs. At runtime, the backend maintained the multi-agent workflow through a custom orchestrator-worker framework coordinating the \textit{ChallengeAgent}, \textit{BuddyAgent}, \textit{CriticAgent}, and \textit{RewardAgent}. LLM inference was served through the Together AI API, with Llama-based models used for agent reasoning and response generation.

Participants accessed \model through a publicly deployed web portal and completed a short interactive session configured at a novice difficulty level. Each session consisted of a small number of cybersecurity challenges, during which participants could request hints, submit responses, and receive automated feedback and reward updates. Sessions were designed to take approximately 5--15 minutes, after which participants completed a post-study survey on engagement, clarity of feedback, perceived learning value, and overall satisfaction.
For analysis, the system recorded challenge prompts, learner responses, hint requests, evaluative feedback, reward assignments, and gameplay state variables such as topic, difficulty level, and XP progression. These logs were used for post-hoc analysis and to support both the human-study findings and the automated evaluation procedures.

\subsection{Human Study}

We conducted a human study to evaluate learner perceptions of \model, focusing on engagement, feedback quality, trust in AI-generated guidance, and perceived learning value. Participants interacted individually with the system and then completed a short post-study survey. We also collected expert feedback from cybersecurity and educational-design evaluators to complement learner responses.

\paragraph{Study Design and Instrument.}
The study used a within-subjects design in which each participant completed one interactive cybersecurity scenario using the full agent loop (\textit{ChallengeAgent} $\rightarrow$ \textit{BuddyAgent} $\rightarrow$ \textit{CriticAgent} $\rightarrow$ \textit{RewardAgent}). The activity exposed participants to progressive challenges, optional scaffolding, automated feedback, and XP-based reward progression. After the interaction, participants completed a brief mixed-format survey containing background questions on prior cybersecurity familiarity, 5-point Likert items assessing learning effectiveness, scenario authenticity, AI hint and feedback quality, trust, challenge fit, adaptivity, gamification impact, engagement, usability, and overall satisfaction, plus one open-ended question for suggested improvements. Participation was voluntary and anonymized.

\paragraph{Expert Feedback.}
To complement learner responses, subject-matter experts in cybersecurity and educational design reviewed the system by completing the same interaction flow. Their feedback focused on scenario realism, challenge calibration, pedagogical alignment, feedback quality, and the motivational value of the reward structure; representative comments are provided in Appendix~\ref{app:expert_feedback}.

\begin{figure}
    \centering
    \includegraphics[width=0.5\linewidth]{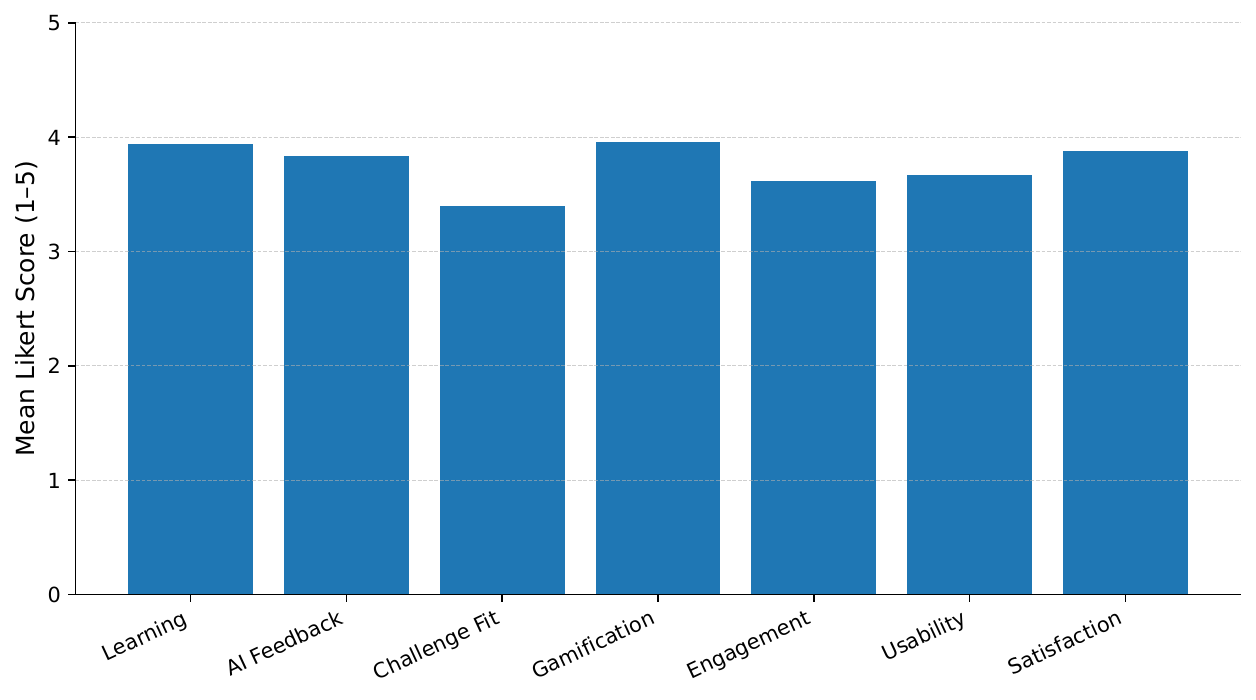}
    \caption{ Mean-Likert ratings aggregated by evaluation category, including learning effectiveness, AI feedback quality, challenge fit, gamification impact, engagement, usability, and overall satisfaction.}
    \label{fig:survey}
\end{figure}

\paragraph{Quantitative Results.}
Likert responses were converted to numerical scores from 1 to 5, and descriptive statistics were computed for each survey item. Overall, participants rated the system positively across most dimensions. The highest-rated dimension was scenario authenticity (M = 4.21, SD = 0.88), followed by gamification impact (M = 3.96, SD = 1.11), trust in system feedback (M = 3.92, SD = 1.18), and willingness to recommend the system for cybersecurity learning (M = 3.88, SD = 0.99). AI-generated hints and explanations were also rated favorably (M = 3.75, SD = 0.99). Engagement (M = 3.62, SD = 1.21), usability (M = 3.67, SD = 1.24), and adaptivity (M = 3.67, SD = 1.31) received moderately positive ratings, while challenge difficulty alignment was lowest (M = 3.12, SD = 1.12), suggesting room for improvement in challenge balancing. Figure~\ref{fig:survey} summarizes the grouped survey results. Cronbach’s alpha across the Likert items was $\alpha = 0.90$, indicating high internal consistency of the survey instrument.

\paragraph{Qualitative Feedback.}
Open-ended responses revealed four recurring themes: trust in AI scaffolding, clarity of feedback, engagement and interaction design, and authenticity of tasks (Table~\ref{tab:themes_clean}). Representative participant quotes are provided in Appendix~\ref{app:user_feedback}. Participants consistently valued the realism of the scenarios and the BuddyAgent’s guided support, often describing the interaction as similar to learning through iterative hints rather than direct answers. At the same time, they requested shorter and more structured responses, clearer transitions between challenge and explanation phases, and more immersive interaction formats. These findings complement the quantitative results by showing that \model was generally perceived as useful, realistic, and trustworthy, while also highlighting opportunities to improve feedback conciseness, interaction flow, and challenge adaptivity.

\begin{table}[t]
\centering
\label{tab:themes_clean}
\small
\setlength{\tabcolsep}{4pt}

\begin{tabularx}{\columnwidth}{>{\raggedright\arraybackslash}p{2.8cm} >{\raggedright\arraybackslash}X}
\rowcolor{gray!20}
\textbf{Theme} & \textbf{Summary and Evidence} \\

\rowcolor{blue!5}
\textbf{AI Scaffolding \& Trust}
& Participants valued the BuddyAgent’s guidance, which supported learning through hints without revealing answers.
\textit{``When it talked me through my responses, it felt like I was really learning.''} \\

\rowcolor{blue!2}
\textbf{Clarity of Feedback}
& Participants requested shorter and more structured explanations. Long responses reduced readability.
\textit{``The paragraphs written by agents are too long. Bullet points would help.''} \\

\rowcolor{blue!5}
\textbf{Engagement \& Interaction}
& Participants suggested improving engagement through more interactive and immersive environments.
\textit{``A terminal-based environment would make the experience more engaging.''} \\

\rowcolor{blue!2}
\textbf{Authenticity of Tasks}
& Challenges were perceived as realistic and aligned with real-world cybersecurity scenarios.
\textit{``Challenges felt genuine and similar to real cybersecurity interviews.''} \\
\end{tabularx}
\caption{Themes identified from open-ended survey responses}
\end{table}

\subsection{LLM-Judge Based Evaluation}

Following human evaluation, we conducted an additional LLM-as-Judge evaluation to further assess the quality of \model interactions at scale. Human evaluation provides direct evidence of perceived system quality, but it is costly to apply across large numbers of transcripts. The LLM-based judge serves as a complementary evaluation layer that enables consistent, rubric-based analysis over many interactions. We evaluate complete transcripts using category-specific prompts that score properties central to the \model design. The evaluation dimensions are: \textit{learner intent alignment, challenge quality, interaction coherence and state tracking, agent role fidelity}, and \textit{domain relevance and cybersecurity grounding}. These metrics directly reflect \model key claims: learner-centered adaptivity, competency-aligned challenge generation, coherent multi-turn orchestration, role-bounded agent behavior, and domain-aware instructional content. For each transcript, the judge assigns a score from 1 to 5 for each category using explicit scoring rubrics and evidence grounded in the transcript. The detailed evaluation dimensions and judge rubrics can be found in Appendix~\ref{app:llm_judge}.

\paragraph{\model results}
The LLM-as-Judge evaluation indicates that \model achieves strong performance across the reported dimensions. The highest scores are obtained in Domain Relevance and Cybersecurity Grounding (4.46) and Agent Role Fidelity (4.35), as seen in Table~\ref{tab:llm_judge_results}, suggesting that the framework consistently maintains domain-focused interactions while preserving the intended division of agent responsibilities. \model also performs well in Challenge Quality and Learner Intent Alignment, indicating that generated challenges are generally well aligned with learner goals and appropriate to the requested topic. Interaction Coherence and State Tracking is somewhat lower (3.95), but still suggests acceptable coherent multi-turn behavior, given that score level 4 in the rubric indicates ``Mostly coherent" behavior (see Appendix~\ref{app:llm_judge}). 

\paragraph{Ablation on \model}
To assess the contribution of expert-informed structural components, we evaluate an ablated version of \model in which the human- and domain-expertise-driven additions, such as the competency map and ontology-based constraints, are removed. Compared with the full system, the ablation yields lower scores on all reported dimensions and the largest degradations appear in challenge quality, learner intent alignment, and domain grounding, indicating that the removed structural components play an important role in maintaining topic alignment, generating appropriate challenges, and preserving cybersecurity-specific relevance. However, the ablation deployment received less traffic than the main \model deployment, leading to fewer interaction transcripts available for evaluation. Therefore, we interpret these results descriptively rather than as directly paired comparisons.


\begin{table}[t]
\centering
\small
\setlength{\tabcolsep}{4pt}
\begin{tabularx}{\columnwidth}{>{\raggedright\arraybackslash}X c c c}
\toprule
\textbf{Eval Dimension} & \textbf{\model} ($n=74$) & \textbf{Ablation} ($n=26$) & $\Delta$ \\
\midrule
Learner Intent Alignment & 4.11 & 3.39 & +0.72 \\
Challenge Quality & 4.20 & 3.46 & +0.74 \\
Interaction Coherence \& State Tracking & 3.95 & 3.54 & +0.41 \\
Agent Role Fidelity & 4.35 & 4.23 & +0.12 \\
Domain Relevance \& Cybersecurity Grounding & 4.46 & 3.77 & +0.69 \\
\bottomrule
\end{tabularx}
\caption{LLM-as-Judge results on the complete \model framework and an ablated version without expert-informed structural components. Scores are averaged on a 1--5 scale; $n$ indicates the number of recorded student interactions. Higher is better.}
\label{tab:llm_judge_results}
\end{table}

\section{Conclusion and Future Work}
This paper presents CyberAgents as a step toward more reliable agentic gamified learning in high-stakes technical domains. By combining competency-based progression, schema-governed behavioral control, and ontology-guided validation, \model shows how adaptive generative interaction can be bounded without sacrificing flexibility or learner support. Beyond the cybersecurity setting, the broader contribution is the concept of \textit{structured autonomy}: a design principle in which agent behavior remains dynamic while being explicitly constrained by pedagogical and domain-level structure.

At the same time, the current study is only an initial step. While the results are encouraging, the relatively small-scale deployment, short interaction duration, and preliminary ablation evidence limit broader claims about long-term learning outcomes. Future work should examine larger evaluations, richer forms of structured autonomy, more immersive gamified interfaces, and the transfer of this framework beyond cybersecurity. More broadly, \model offers both a practical starting point and a conceptual blueprint for reliable, pedagogically aligned agentic learning systems.


\section*{Ethics Statement}

This work investigates the use of agentic AI for gamified cybersecurity learning, a setting that raises both educational and domain-specific ethical considerations. Because cybersecurity is a high-stakes domain, inaccurate or unsafe generated content could mislead learners or inadvertently expose them to harmful procedural guidance. To mitigate these risks, CyberAgents incorporates ontology-guided validation, schema-governed behavioral control, and pre-display safety checks designed to constrain domain reasoning, preserve pedagogical alignment, and reduce unsafe instructional drift.

The system is intended strictly for educational use and is designed to support conceptual understanding rather than operational misuse. Challenges, hints, and feedback are grounded in curriculum-derived competencies and validated against structured cybersecurity constraints before being shown to learners. We do not position the system as a substitute for expert instruction, and we recognize that AI-generated feedback may still be imperfect despite these safeguards. Human oversight therefore remains important, particularly in high-stakes instructional settings.

The human study was conducted with informed and voluntary participation. Responses were anonymized, no identifying personal information was required, and collected interaction traces were used only for research and evaluation purposes. Participants evaluated the system in a controlled educational setting, and the study focused on perceptions of learning, engagement, trust, and feedback quality rather than making claims about credentialing or high-stakes assessment.

More broadly, this work takes the position that educational agentic AI should be developed with explicit attention to reliability, safety, transparency, and pedagogical responsibility. We view structured autonomy not only as a technical design principle, but also as an ethical one: constraining agent behavior and domain reasoning is essential for building trustworthy learning systems in sensitive technical domains such as cybersecurity.

\section{Acknowledgments}
This work is supported by the National Science Foundation (NSF) under grant SaTC (\#2335666). Any opinions, findings, conclusions, or recommendations expressed in this material are those of the author(s) and do not necessarily reflect the views of the National Science Foundation.

\bibliography{colm2026_conference}
\bibliographystyle{colm2026_conference}

\appendix

\section{Learner and Expert Perspectives}
\label{app:user_feedback}

\subsection{User Feedback}
\label{app:user_feedback}
\begin{tcolorbox}[
    colback=gray!4,
    colframe=gray!60,
    title={Representative Participant Quotes},
    fonttitle=\bfseries,
    boxrule=0.5pt,
    arc=1.5pt
]
\textbf{Trust in AI Guidance.}
\textbf{P1:} ``When it talked me through my responses and provided hints, it felt like I was really learning.''  
\textbf{P2:} ``I like how the BuddyAgent engages conversationally and the color changes between agents.''

\vspace{4pt}
\textbf{Clarity of Feedback and Interaction.}
\textbf{P3:} ``The paragraphs written by agents are too long. It would be better in bullet points.''  
\textbf{P4:} ``I expected the system to explain the topic, but it started giving challenges instead.''

\vspace{4pt}
\textbf{Authenticity and Engagement.}
\textbf{P5:} ``Challenges felt genuine and similar to what I would expect in a cybersecurity interview.''  
\textbf{P6:} ``A terminal-based environment would make the experience more engaging and technical.''

\vspace{4pt}
\textbf{Topic Structure and Navigation.}
\textbf{P7:} ``It would help to start with a map of cybersecurity topics and refer back to it.''
\end{tcolorbox}

\subsection{Domain Expert Feedback}
\label{app:expert_feedback}

\begin{tcolorbox}[
    colback=blue!3,
    colframe=blue!45!black,
    title={Cybersecurity Expert Feedback},
    fonttitle=\bfseries,
    boxrule=0.6pt,
    arc=2pt,
    left=6pt,
    right=6pt,
    top=6pt,
    bottom=6pt
]
The cybersecurity expert found the multi-agent design pedagogically effective, noting that the coordination among the BuddyAgent, ChallengeAgent, CriticAgent, and RewardAgent creates a realistic tutor--practice--feedback loop while reducing the burden on any single model. They also highlighted the technical accuracy of the cybersecurity content, including correct use of the Principle of Least Privilege in firewall configuration and sound explanations of Cross-Site Scripting defenses through input validation and sanitization. The BuddyAgent’s structured conversational cues were viewed as especially helpful for novice learners, as they lower the barrier to entry by presenting guided options with relevant industry terminology. In addition, the expert noted that the gamified feedback and reward signals create an encouraging learning environment that supports motivation.

At the same time, the expert identified two areas for refinement: occasional topical shifts between agents suggest a need for stronger state sharing to ensure smoother curricular continuity, and some BuddyAgent confirmations could be made more concise and conversational to improve the naturalness of the tutoring experience.
\end{tcolorbox}

\section{LLM-Judge based Evaluation}
\label{app:llm_judge}

Following the human evaluation, we conduct an additional LLM-as-Judge evaluation to further assess the quality of CyberAgents interactions at scale. Detailed evaluation dimensions and prompts used are as below:

\begin{promptbox}
\textbf{Universal Judge Instructions}

You are an expert evaluator of AI learning systems, educational dialogue, and multi-agent orchestration.

You are evaluating one conversation transcript from a gamified cybersecurity learning system.

\begin{itemize}[leftmargin=1.2em, nosep]
    \item Score only what is explicitly observable in the transcript.
    \item Do not reward politeness, verbosity, or technical-sounding language unless they improve the target dimension.
    \item If a relevant agent or behavior does not appear in the transcript, mark that subcriterion as \textit{not observed} instead of penalizing it unfairly.
    \item Ground reasoning in concrete transcript evidence.
    \item Keep justifications concise and evidence-based.
\end{itemize}

Score labels: 1 = very poor, 2 = poor, 3 = mixed, 4 = good, 5 = excellent.
\end{promptbox}

\begin{promptbox}
\textbf{Learner Intent Alignment}

Evaluate whether the system follows the learner's explicitly stated goals, including requested topic and experience level.

\textbf{Subcriteria}
\begin{itemize}[leftmargin=1.2em, nosep]
    \item \textbf{Topic alignment}: Does the system stay on the learner's requested subject?
    \item \textbf{Level alignment}: Are explanations or challenges appropriate for the learner's stated level?
\end{itemize}

\textbf{Scoring guidance}
\begin{itemize}[leftmargin=1.2em, nosep]
    \item 1: Repeatedly ignores explicit learner intent
    \item 2: Acknowledges intent but frequently misaligns
    \item 3: Mixed; some alignment but notable failures
    \item 4: Mostly aligned with minor lapses
    \item 5: Strongly and consistently aligned
\end{itemize}
\end{promptbox}

\begin{promptbox}
\textbf{Challenge Quality}

Evaluate how well the system generates learning activities or tasks for the learner.

\textbf{Subcriteria}
\begin{itemize}[leftmargin=1.2em, nosep]
    \item \textbf{Challenge-topic relevance}: Is the challenge aligned with the learner's requested topic?
    \item \textbf{Difficulty appropriateness}: Is the challenge suitable for the learner's stated skill level?
    \item \textbf{Objective clarity}: Is the task clearly defined with understandable goals?
\end{itemize}

\textbf{Scoring guidance}
\begin{itemize}[leftmargin=1.2em, nosep]
    \item 1: Clearly off-topic, poorly calibrated, or unusable
    \item 2: Major weaknesses in relevance or clarity
    \item 3: Mixed quality
    \item 4: Good challenge design
    \item 5: Highly relevant, clear, and pedagogically strong challenge
\end{itemize}
\end{promptbox}

\begin{promptbox}
\textbf{Interaction Coherence and State Tracking}

Evaluate whether the interaction remains coherent across turns and consistent with prior context.

\textbf{Subcriteria}
\begin{itemize}[leftmargin=1.2em]
    \item \textbf{Local coherence}: Does each response make sense given the immediately preceding turn?
    \item \textbf{Cross-turn consistency}: Does the system remain consistent across the conversation?
\end{itemize}

\textbf{Scoring guidance}
\begin{itemize}[leftmargin=1.2em]
    \item 1: Severe incoherence or state failure
    \item 2: Frequent continuity failures
    \item 3: Mixed coherence
    \item 4: Mostly coherent
    \item 5: Strongly coherent and state-aware
\end{itemize}
\end{promptbox}

\begin{promptbox}
\textbf{Domain Relevance and Cybersecurity Grounding}

Evaluate whether the system remains grounded in the requested cybersecurity topic and uses domain concepts appropriately.

\textbf{Subcriteria}
\begin{itemize}[leftmargin=1.2em]
    \item \textbf{Cybersecurity topic relevance}: Stays focused on the learner's requested cybersecurity topic
    \item \textbf{High-level concept correctness}: Uses cybersecurity concepts in a broadly correct way
    \item \textbf{Example-to-concept alignment}: Uses examples that match the intended concept
\end{itemize}

\textbf{Scoring guidance}
\begin{itemize}[leftmargin=1.2em]
    \item 1: Major domain mismatch or confusion
    \item 2: Weak grounding with substantial irrelevant content
    \item 3: Mixed relevance and grounding
    \item 4: Mostly well-grounded
    \item 5: Strong, accurate, topic-aligned grounding
\end{itemize}
\end{promptbox}
\begin{promptbox}
\textbf{Agent Role Fidelity}

Evaluate whether each agent behaves according to its intended instructional role.

\textbf{Expected roles}
\begin{itemize}[leftmargin=1.2em]
    \item \textbf{BuddyAgent}: support, hints, guidance, redirection
    \item \textbf{ChallengeAgent}: generate tasks/challenges
    \item \textbf{CriticAgent}: evaluate learner responses and provide critique
    \item \textbf{RewardAgent}: assign XP, badges, or progression rewards
\end{itemize}

\textbf{Subcriteria}
\begin{itemize}[leftmargin=1.2em]
    \item BuddyAgent role fidelity
    \item ChallengeAgent role fidelity
    \item CriticAgent role fidelity
    \item RewardAgent role fidelity
    \item Role sequencing
\end{itemize}

\textbf{Scoring guidance}
\begin{itemize}[leftmargin=1.2em]
    \item 1: Severe role confusion
    \item 2: Multiple important violations
    \item 3: Mixed role adherence
    \item 4: Mostly correct role behavior
    \item 5: Clear and consistent role fidelity
\end{itemize}
\end{promptbox}

\clearpage
\section{Representative Model Artifacts: Ontology, Schemas, and Prompts}
\label{app:model_schema}
\subsection{CyberSecurity Ontology and Competency Mapping}
\begin{promptbox}
\textbf{AISecKG Ontology Sample Edges}
\begin{verbatim}
('user', 'uses', 'app'), 
('feature', 'can_expose', 'vulnerability'),
('securityTeam', 'can_detect', 'vulnerability'),
...
\end{verbatim}
\end{promptbox}

\begin{promptbox}
\textbf{Sample of Cybersecurity Competency Map Elements}
\begin{verbatim}
{
  "attacks": {
    "Man-in-the-Middle (MitM) Attack": {
      "description": "Intercepting and altering communication between 
        two parties.",
      "difficulty_level": 3,
      "exploits_vulnerabilities": [
        "weak_encryption"
      ],
      "gaming_appeal": "medium",
      "id": "a4",
      "impact_level": "high",
      "requires_tools": [
        "Wireshark"
      ]
  },
  "tools": {
    "Wireshark": {
      "category": "analysis",
      "description": "A network protocol analyzer used for network 
        troubleshooting and analysis.",
      "difficulty_level": 2,
      "gaming_appeal": "high",
      "id": "t1",
      "related_concepts": [
        "network_protocols",
        "TCP_handshake"
      ]
    },
}
\end{verbatim}
\end{promptbox}

\begin{promptbox}
\textbf{Enhanced Ontology Schema}
\begin{verbatim}
Enhanced Cybersecurity Competency Ontology

**Core Entity Types:**
1. Tool: {id, name, description, difficulty_level, category}
   - Examples: nmap, wireshark, metasploit

2. Technique: {id, name, description, difficulty_level, attack_phase}
   - Examples: port_scanning, vulnerability_exploitation, social_engineering

3. Concept: {id, name, description, difficulty_level, domain}
   - Examples: TCP_handshake, network_protocols, encryption

4. Attack: {id, name, description, difficulty_level, impact_level}
   - Examples: DoS_attack, SQL_injection, buffer_overflow

5. Vulnerability: {id, name, description, severity, exploitability}
   - Examples: unpatched_software, weak_passwords, misconfiguration

6. Defense: {id, name, description, difficulty_level, effectiveness}
   - Examples: firewall_configuration, intrusion_detection, incident_response

**Competency Levels:**
- Level 1 (Beginner): Basic concepts, terminology, simple tools
- Level 2 (Intermediate): Tool usage, basic techniques, understanding 
relationships
- Level 3 (Advanced): Complex techniques, tool combination, analysis skills
- Level 4 (Expert): Advanced attacks, sophisticated defenses, strategic thinking

**Gaming-Oriented Relationships:**
- PREREQUISITE: Concept A must be learned before Concept B
- ENABLES: Learning A enables the use of B
- COUNTERS: Defense A counters Attack B
- EXPLOITS: Attack A exploits Vulnerability B
- DETECTS: Tool A can detect Attack B
- CLUSTERS_WITH: Concepts that should be taught together

\end{verbatim}
\end{promptbox}

\subsection{Game Flow and System Prompts}
\begin{promptbox}
\textbf{System Prompt: AssemblerAgent}
\begin{verbatim}
You are a prompt generation module for a multi-agent learning system. Your job is 
to generate clear, role-specific prompts that guide each AI agent to behave 
correctly based on their metadata.

Each agent receives a properties object that contains key configuration fields. 
Your generated prompt must:
1. Clearly define the agent's role and responsibility.
2. Explicitly instruct the agent how to use each property.
3. Constrain the agent's behavior based on these properties.
4. Ensure the output format is aligned with the use case 
    (e.g., challenge, hint, feedback, reward).

You will be given 
- `agent_name`: the name of the agent (e.g., "ChallengeAgent", "BuddyAgent")
- `agent_role`: a description of what this agent is supposed to do
- `property_schema`: a JSON schema or list of fields relevant to this agent

Keep in mind that you are intializing the properties of the agent using the 
schema information, so make sure to decide on a value for all fields.

ONLY output the final prompt text, no extra commentary.

\end{verbatim}
\end{promptbox}

\subsection{Representative Behavioral Schemas for Agent Control}

\begin{tcolorbox}[
    colback=gray!4,
    colframe=gray!60,
    title={Representative Agent-Schema Metadata},
    fonttitle=\bfseries,
    boxrule=0.5pt,
    arc=1.5pt,
    breakable
]
\textbf{ChallengeAgent}
\begin{verbatim}
{
  "Mode": "solo challenge",
  "Topic": "SQL Injection",
  "Level": "beginner",
  "StudentLearnerProfile": "undergrad",
  "Objective": "Identify and explain how an attacker could exploit a vulnerable 
    web application using SQL Injection, and propose a basic mitigation 
    strategy.",
  "ChallengeType": "problem-solving puzzle"
}
\end{verbatim}

\textbf{BuddyAgent}
\begin{verbatim}
{
  "Mode": "support_proactive",
  "HintType": "strategic",
  "SupportLevel": "high",
  "ResponseStyle": "friendly",
  "TriggerCondition": "performance_based"
}
\end{verbatim}

\textbf{CriticAgent}
\begin{verbatim}
{
  "Mode": "real_time",
  "Criteria": ["correctness", "completeness", "efficiency"],
  "FeedbackStyle": "constructive"
}
\end{verbatim}

\textbf{RewardAgent}
\begin{verbatim}
{
  "Mode": "reward",
  "RewardType": "points",
  "RewardStyle": "celebratory",
  "ProgressSignal": "progress_bar",
  "Rubric": {
    "Tier": ["Legendary", "Expert", "Practitioner", "Explorer", "Apprentice"],
    "XP_Range": ["90-100", "75-89", "60-74", "45-59", "0-44"],
    "Badge": ["Gold", "Silver", "Bronze", "Explorer", "Encouragement"],
    "ScoringGuidelines": {
      "Base_XP": 50,
      "Autonomy_Bonus": "+10-30",
      "Feedback_Integration_Bonus": "+10-20",
      "Accuracy_Bonus": "+10",
      "Over_Scaffolding_Penalty": "-10-20",
      "XP_Cap": 100
    }
  }
}
\end{verbatim}
\end{tcolorbox}

\begin{figure*}[t]
    \centering
    \includegraphics[width=1\textwidth]{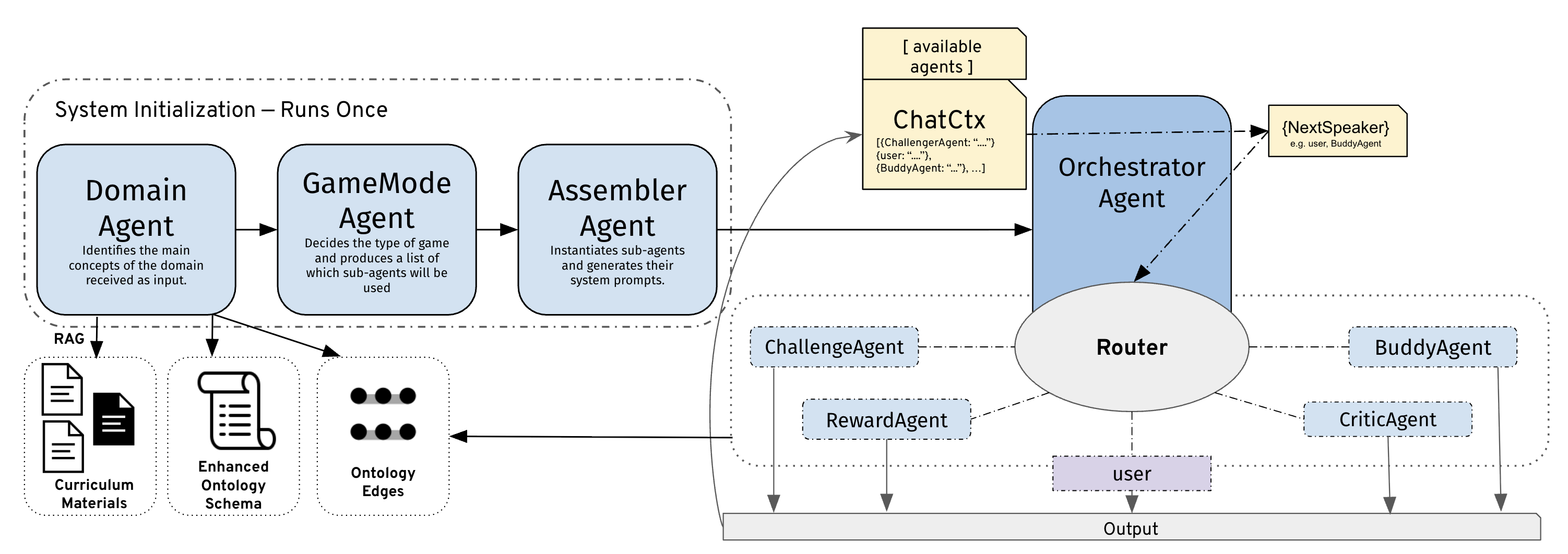}
    \caption{\textbf{CyberAgents data-flow architecture.} \textit{DomainAgent} performs retrieval-augmented processing over curriculum materials and synthesizes a competency map from the enhanced ontology schema and ontology edges. This competency map is then provided to \textit{GameModeAgent} and \textit{AssemblerAgent}, which dynamically generate system prompts for the worker agents. \textit{OrchestratorAgent} oversees the interaction by tracking conversation state and determining speaker flow. The worker agents---\textit{ChallengeAgent}, \textit{BuddyAgent}, \textit{CriticAgent}, and \textit{RewardAgent}---access the ontology and carry out their respective roles when selected by the \textit{OrchestratorAgent}.}
    \label{fig:appendix_ss1}
\end{figure*}

\begin{figure*}[t]
    \centering
    \includegraphics[width=1\textwidth]{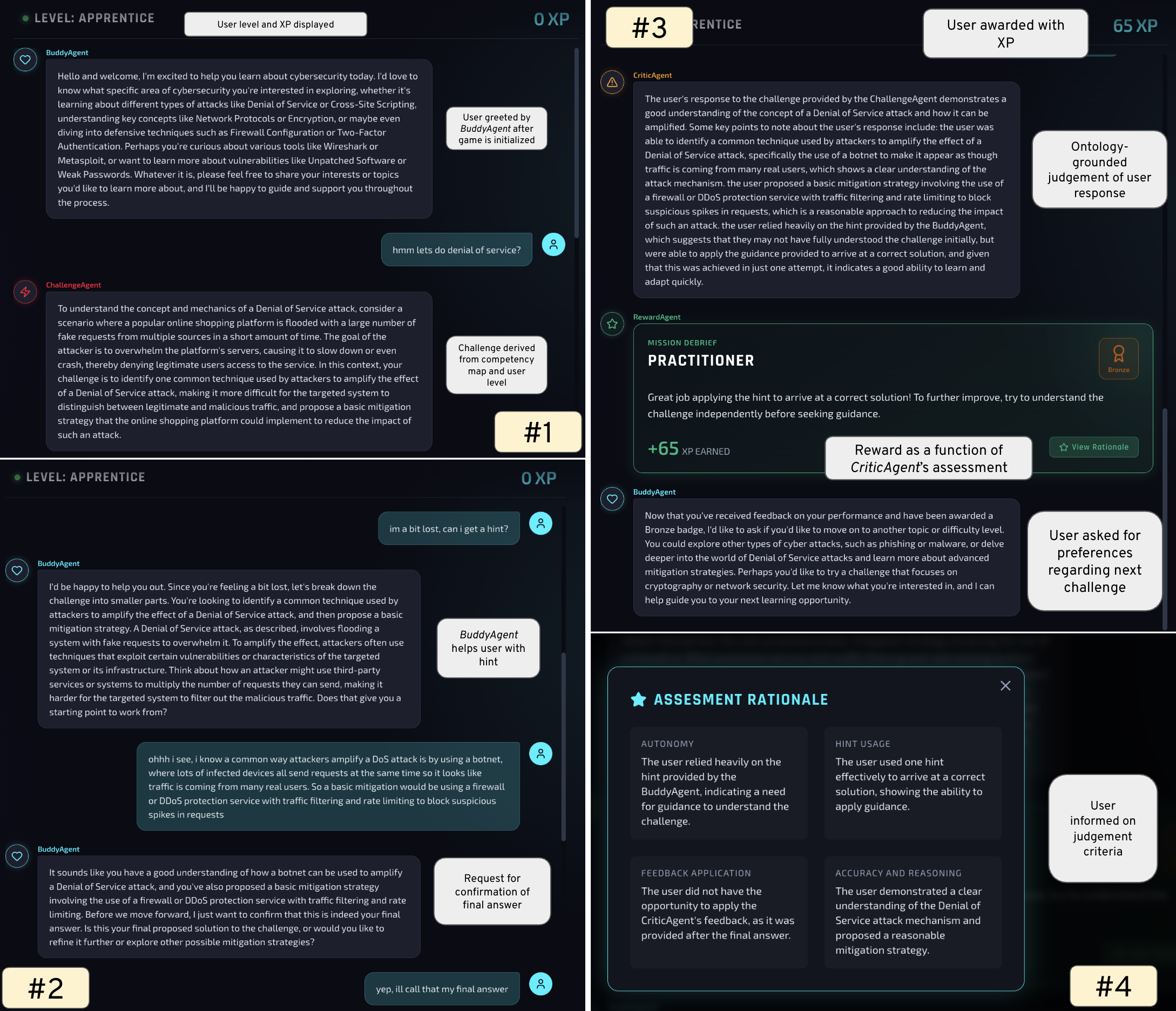}
    \caption{\textbf{CyberAgents interaction flow.} Four representative snapshots of a learner session: (\#1) session initialization and challenge setup, (\#2) BuddyAgent hinting and final-answer confirmation, (\#3) CriticAgent evaluation with RewardAgent XP assignment and next-challenge prompting, and (\#4) learner-facing assessment rationale.}
    \label{fig:cyberagents_interaction_flow}
\end{figure*}

\end{document}